\documentclass[conference]{IEEEtran}
\IEEEoverridecommandlockouts

\usepackage{times}
\usepackage[T1]{fontenc}
\usepackage{amsmath,amssymb,amsfonts}
\usepackage{graphicx}
\usepackage{booktabs}
\usepackage{multirow}
\usepackage{xcolor}
\usepackage{url}
\usepackage{algorithm}
\usepackage{algpseudocode}
\usepackage{caption}
\usepackage{subcaption}

\newcommand{\R}{\mathbb{R}}
\newcommand{\Tcal}{\mathcal{T}}
\newcommand{\Rcal}{\mathcal{R}}
\newcommand{\Ical}{\mathcal{I}}
\newcommand{\Ecal}{\mathcal{E}}
\newcommand{\Gcal}{\mathcal{G}}
\newcommand{\Mcal}{\mathcal{M}}
\newcommand{\Kcal}{\mathbb{K}}
\newcommand{\eps}{\varepsilon}

\title{Signal2Symbol: Neuro-Symbolic Temporal Reasoning for Explainable Physiological Time-Series Anomaly Detection}
\DeclareRobustCommand{\IEEEauthorrefmark}[1]{\smash{\textsuperscript{\footnotesize #1}}}
\author{
  \IEEEauthorblockN{
      Naser Mansour\IEEEauthorrefmark{1}, 
        Sidahmed Benabderrahmane\IEEEauthorrefmark{1}\IEEEauthorrefmark{*}, 
        Ameer Rahwan\IEEEauthorrefmark{2}
 }

  \IEEEauthorblockA{
  \IEEEauthorrefmark{1} 
  New York University, NYUAD, Computer Science Department, Division of Science.} 
    \IEEEauthorblockA{\IEEEauthorrefmark{2} Koc University, Faculty of Medicine, Turkey.}    
    \IEEEauthorblockA{ \{nmd9736@nyu.edu; sidahmed.benabderrahmane@gmail.com; arahwan24@ku.edu.tr\}} 
}
\begin{document}
\maketitle

% ============================================================
\begin{abstract}
Physiological time series such as electrocardiograms (ECG) and electroencephalograms (EEG)
exhibit complex temporal structure, substantial acquisition variability, and a strong need for
transparent decision-making.
Although deep models can achieve high detection performance, they
often provide limited insight into \emph{why} a segment is anomalous, \emph{how} local anomalies
relate over time, and \emph{whether} a detection belongs to a broader recurring pattern.
We propose \textbf{Signal2Symbol}, a neuro-symbolic framework for explainable biosignal anomaly
detection.
The method first converts ECG/EEG signals into symbolic sequences using either a
learned VQ-VAE (Vector Quantized Variational Autoencoder) codebook or a SAX (Symbolic Aggregate approXimation) baseline.
It then constructs bigram-enriched token-window transactions and scores anomalies through
rare itemset evidence derived from minimal rare itemset mining.
Detected anomalous windows are merged into intervals and related using Allen interval algebra,
enabling composite temporal explanations such as escalation chains, artifact overlap, and
cross-channel synchrony.
Finally, we introduce a \textbf{rare temporal concept lattice} based on Formal Concept Analysis
(FCA), which groups anomalous intervals by shared rare symbolic evidence, Allen temporal
relations, channel context, and robustness attributes.
The resulting Galois lattice compresses many local detections into interpretable families of
temporal-symbolic anomalies.
We evaluate on three public benchmarks---MIT-BIH Arrhythmia (beat-level ECG), PTB-XL
(record-level ECG), and the Bonn EEG dataset (segment-level EEG)---and stress-test robustness under additive noise and baseline-wander perturbations. The results highlight the value of
neuro-symbolic tokenization for temporal anomaly analysis and show that Allen/FCA reasoning
provides compact, interpretable summaries of local detections.
\end{abstract}


% ============================================================
\section{Introduction}

Physiological time-series anomaly detection is central to e-health, clinical monitoring,
and biomedical decision support \cite{Darban24,Liu24}.
Signals such as ECG and EEG contain subtle temporal structures that may indicate arrhythmias,
ischemic changes, seizures, abnormal brain dynamics, or acquisition artifacts \cite{Tabassum24,Niu25}.
In practice, physiological signals are affected by sensor noise, baseline wander, device
differences, missing channels, and population shift.
A useful anomaly detector must therefore be robust to drift while also providing interpretable
evidence.

Most modern machine learning methods output a numerical anomaly score or a saliency heatmap.
Such outputs may be useful for ranking, but they often fail to answer three questions that
matter to analysts and clinicians:
\emph{What symbolic pattern made this segment unusual?}
\emph{How does this anomaly relate to nearby events over time?}
\emph{Does this interval belong to a recurring family of anomalies?}

\textbf{Key idea.}
We convert continuous physiological signals into symbolic sequences, detect rare symbolic
co-occurrences, lift local detections into temporal intervals, and organize the resulting
intervals into interpretable concept lattices.
The pipeline follows:
%\[
%\text{signal} \rightarrow \text{tokens} \rightarrow \text{transactions}
%\rightarrow \text{rare patterns} \rightarrow \text{intervals} 
%\rightarrow \text{Allen relations} \rightarrow \text{FCA concepts}.
%\]
\[
\begin{aligned}
\text{signal} &\rightarrow \text{tokens} \rightarrow \text{transactions} \rightarrow \text{rare patterns}\\
& \rightarrow \text{intervals} \rightarrow \text{Allen relations} \rightarrow \text{FCA concepts}.
\end{aligned}
\]
This design combines neural representation learning, symbolic rarity mining, temporal
reasoning, and concept-based explanation.

\textbf{Contributions.}
This paper makes four main contributions:
\begin{itemize}
  \item \textbf{Symbolization for biosignals:} We compare learned \textbf{VQ-VAE tokenization}
  against \textbf{SAX} as a transparent symbolic baseline for ECG/EEG signals.
  \item \textbf{Rare-pattern anomaly scoring:} We propose a rarity-based scoring function
  that aggregates evidence from minimal rare itemsets over bigram-enriched sliding symbolic
  windows, and connect the weighting to an information-theoretic surprisal interpretation.
  \item \textbf{Temporal interval reasoning:} We transform high-scoring windows into anomaly
  intervals and use Allen interval algebra to detect temporal configurations such as
  escalation, artifact overlap, staged onset, and cross-channel synchrony.
  \item \textbf{Rare temporal concept lattices:} We introduce an FCA/Galois-lattice layer
  that groups anomalous intervals according to shared rare motifs, Allen relations, channel
  context, reconstruction fidelity, and robustness attributes, yielding compact visual
  explanations.
\end{itemize}

% ============================================================
\section{Related Work}

\textbf{Symbolic representations of time series.}
Symbolic Aggregate approXimation (SAX)~\cite{sax} and related discretization methods transform
continuous time series into symbolic strings, enabling motif discovery, rule mining, indexing,
and interpretable analysis \cite{jobts27}.
SAX is attractive because it is simple, fast, and transparent, but its quantization is based
mainly on amplitude discretization and may miss morphology-level differences in ECG or EEG
waveforms \cite{simts14}.

\textbf{Discrete neural tokenizers.}
Vector-Quantized VAEs (VQ-VAE)~\cite{vqvae} learn a discrete codebook in a latent space.
Each signal window is encoded, assigned to the nearest codebook vector, and decoded for
reconstruction.
The code index can be interpreted as a learned symbol.
Compared with fixed discretization, VQ-VAE learn prototypes to capture recurring
waveform shapes, for a suitable biosignal tokenization.

\textbf{Anomaly detection and explainability in biosignals.}
ECG/EEG anomaly detection spans feature-based models, density estimators, isolation-based
methods, autoencoders, CNNs, transformers, and hybrid pipelines \cite{Arulmurugan26,Darban24,Liu24,Tabassum24,Niu25}.
Explainability is often provided through saliency maps or reconstruction errors.
In contrast, symbolic approaches can provide direct evidence in the form of motifs,
co-occurrences, and rules \cite{Liang26}.

\textbf{Temporal reasoning.}
Allen interval algebra~\cite{allen1983} defines qualitative relations between time intervals,
including \texttt{before}, \texttt{meets}, \texttt{overlaps}, \texttt{starts}, \texttt{during},
and \texttt{finishes}.
In clinical monitoring, interval-level reasoning helps convert local detections into
higher-level temporal patterns such as escalation, repeated bursts, or artifact-contaminated
detections.

\textbf{Formal Concept Analysis.}
Formal Concept Analysis (FCA)~\cite{ganter1999} derives a concept lattice from a
binary relation between objects and attributes.
Each formal concept groups a set of objects sharing a maximal set of attributes.
FCA has been used for interpretable knowledge discovery because it organizes patterns in a
lattice structure rather than as an unstructured list of rules.
In this work, we use FCA as an explanation layer over detected anomaly intervals: objects are
anomalies, and attributes are rare symbolic evidence, Allen relations, and signal-context
predicates.

% ============================================================
\section{Problem Formulation and Overview}

\subsection{Problem Statement}

Let $x(t)\in \R^{C}$ denote a multichannel physiological time series of length $T$ with $C$
channels, corresponding to ECG leads or EEG sensors.
We segment $x(t)$ into $N$ units:
\[
\{(x_i,[t_i^s,t_i^e])\}_{i=1}^{N},
\]
where each $x_i$ is either an ECG beat-centered window or a fixed-length EEG window.

The objective is to produce:
\begin{enumerate}
  \item a record-level anomaly score $S_{\text{rec}}(x)$;
  \item a sequence of window-level scores $\{S_{\text{win}}(w_j)\}$;
  \item anomaly intervals $\Ical(x)=\{(a_j^s,a_j^e,\text{conf}_j)\}$;
  \item local explanations through rare symbolic patterns;
  \item temporal explanations through Allen relations;
  \item concept-level explanations through an FCA lattice.
\end{enumerate}

\subsection{Pipeline Overview}

\begin{figure}[t]
\centering
\includegraphics[width=\linewidth]{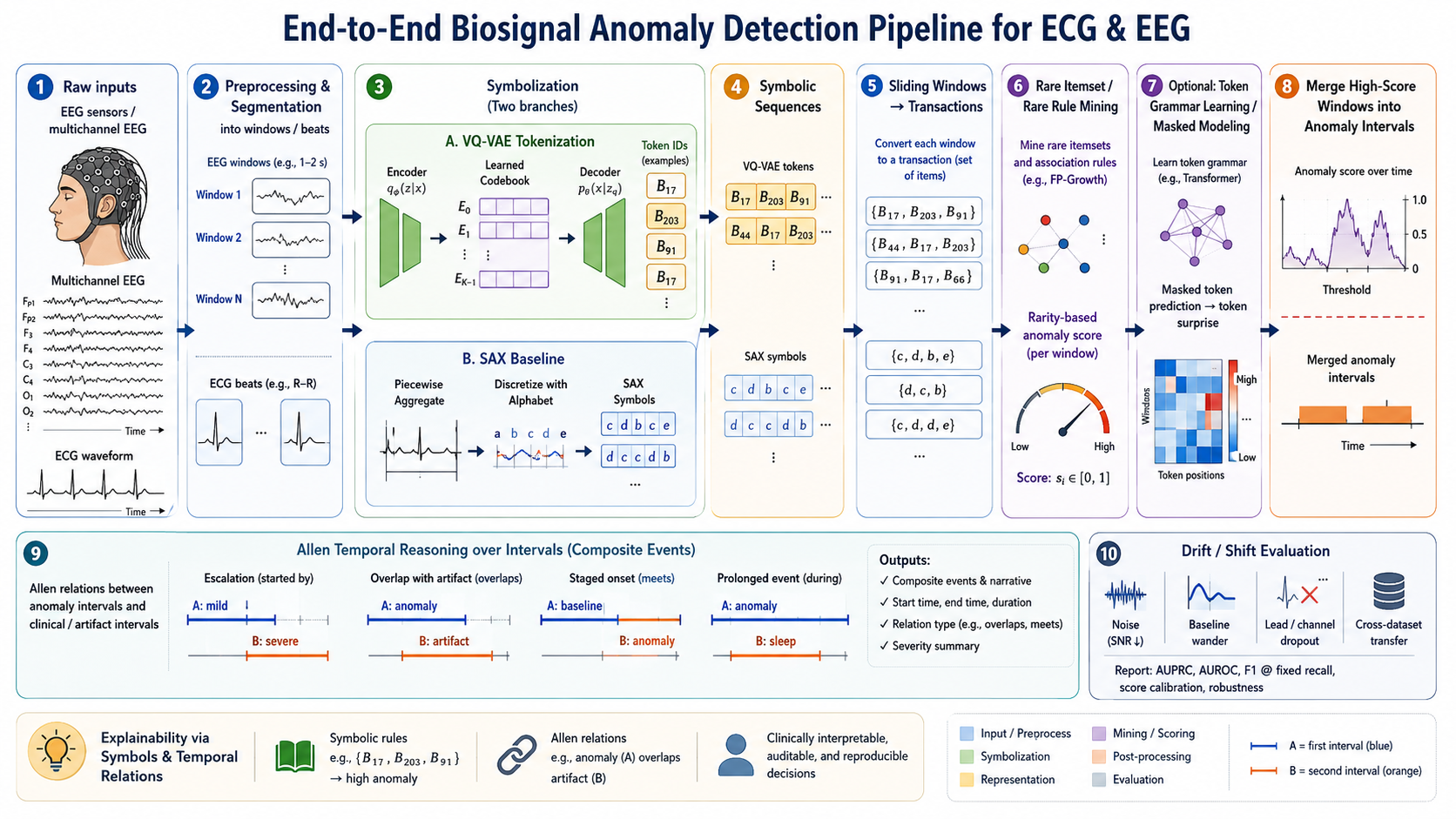}
\caption{Signal2Symbol pipeline.
Raw ECG/EEG signals are tokenized using SAX or VQ-VAE, transformed into bigram-enriched
symbolic transactions, scored by rare-pattern evidence, merged into intervals, related through
Allen algebra, and organized into FCA/Galois lattices for visual explanation.}
\label{fig:pipeline_overview}
\vspace{-1.5 em}
\end{figure}

Figure~\ref{fig:pipeline_overview} summarizes the complete pipeline.
Raw signals are first symbolized using SAX or VQ-VAE.
The resulting token stream is converted into bigram-enriched sliding-window transactions.
Rare itemsets define a symbolic anomaly score.
High-scoring windows are merged into anomaly intervals.
Allen relations are computed between intervals.
Finally, FCA organizes anomalous intervals into a rare temporal concept lattice.

% ============================================================
\section{Symbolization of Time Series}

\subsection{SAX Baseline}

Given a univariate series $y(t)$, SAX~\cite{sax} applies z-normalization, Piecewise Aggregate
Approximation (PAA), and discretization into an alphabet of size $A$.
For multichannel data, we apply SAX per channel or per selected channel group and concatenate
the resulting symbols.
SAX provides a simple and transparent baseline for symbolic anomaly detection.

\textbf{Hyperparameters.}
We vary alphabet size $A\in\{4,5,6\}$ and PAA resolution.
These parameters control the trade-off between coarse robustness and fine symbolic resolution.

\subsection{VQ-VAE Tokenization}

VQ-VAE~\cite{vqvae} learns a dictionary of prototypical signal shapes.
For each segment $x_i$, the encoder produces a latent vector: $h_i = E_\theta(x_i)\in \R^d$. The vector is mapped to the nearest codebook entry: $z_i = \arg\min_{k\in\{1,\ldots,K\}}\|h_i-e_k\|_2$, where $\{e_k\}_{k=1}^{K}$ is the learned codebook.
The selected entry is decoded: $\hat{x}_i = D_\phi(e_{z_i})$. The discrete index $z_i$ is converted into a categorical symbolic token: $B_{z_i}$ for an ECG beat-centered segment and $E_{z_i}$ for an EEG window. Thus, an ECG segment assigned to codebook entry $17$ becomes \texttt{B\_17}, while an EEG segment assigned to entry $42$ becomes \texttt{E\_42}. Because each token inherits the timestamp interval $[t_i^s,t_i^e]$ of its source segment, the tokenizer outputs a time-aligned symbolic sequence.

\textbf{Training.}
We train VQ-VAE exclusively on \emph{normal} signal segments.
The codebook therefore learns prototypes of typical waveform morphologies.
When an anomalous segment is presented at test time, it is mapped to its nearest normal
prototype; the anomaly signal arises from unusual \emph{combinations} of these prototypes
across sliding windows.
This relational design means that codebook granularity and utilization directly affect
sensitivity: dead codes reduce effective resolution, motivating the use of exponential
moving average (EMA) codebook updates and codebook utilization monitoring.

\textbf{Reconstruction fidelity.}
We also record the per-segment reconstruction error
$r_i = \|x_i - \hat{x}_i\|_2$, which provides a complementary anomaly signal.
Segments that are poorly represented by any codebook prototype yield high reconstruction
error and are flagged via a dedicated FCA attribute (Section~\ref{sec:fca}).

\textbf{Segmentation.}
ECG signals are represented using either R-peak-centered beat windows or complete record-level segments, depending on the evaluation granularity. EEG signals are divided into fixed-length windows with overlap.

\textbf{Ablations.}
We study codebook sizes $K\in\{128,256,512\}$ and analyze the impact of dead codes,
reconstruction quality, and token entropy.

\begin{figure}[t]
\centering
\includegraphics[width=1\linewidth]{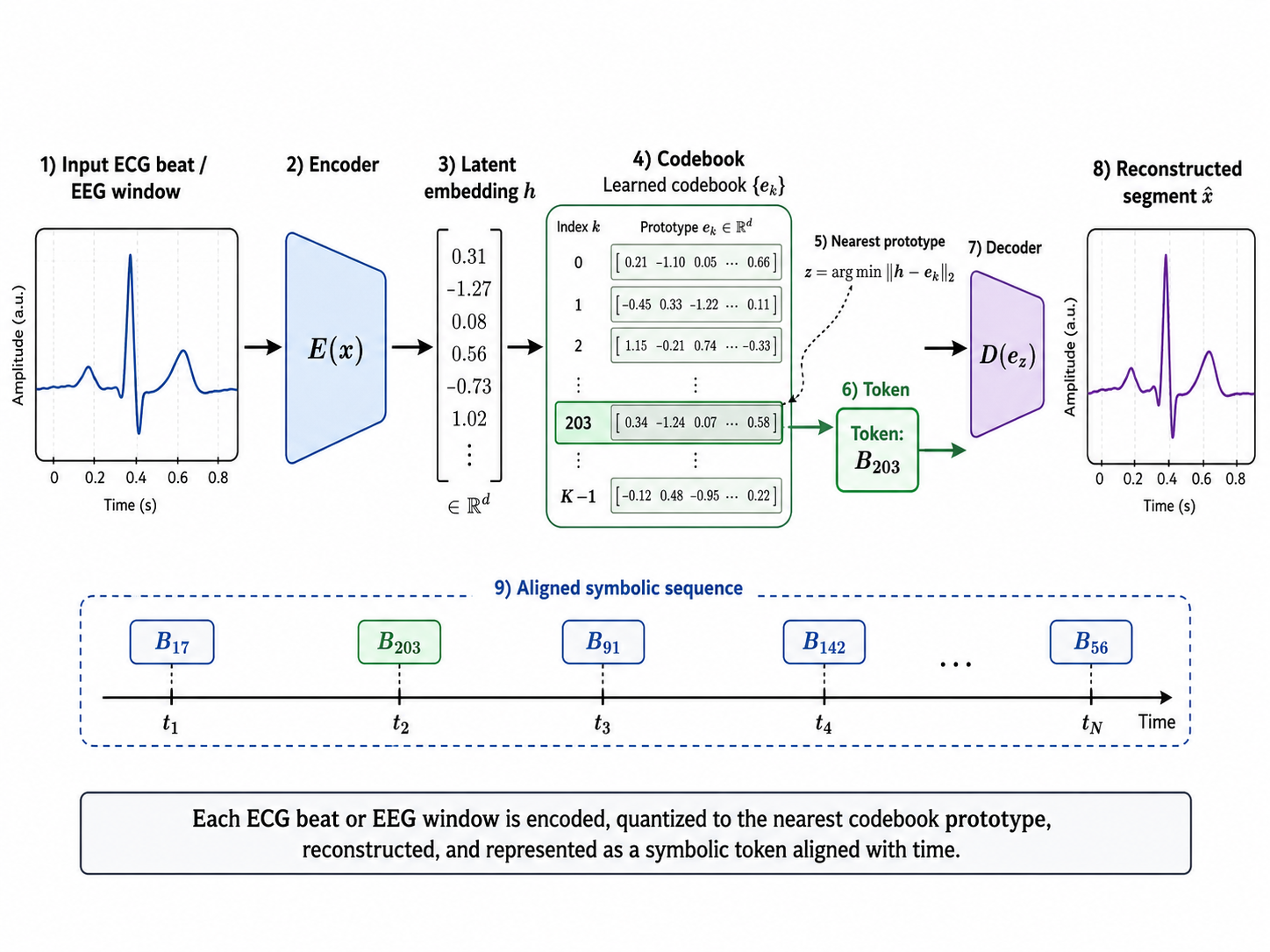}
\vspace{-2em}
\caption{VQ-VAE tokenization.
Each ECG beat or EEG window is encoded, assigned to the nearest codebook prototype,
reconstructed through a decoder, and represented as a symbolic token aligned with timestamps.}
\label{fig:vqvae}
\vspace{-1.7em}

\end{figure}
The output of this section is a timestamped symbolic sequence:
\[
S(x)=\{(s_i,[t_i^s,t_i^e])\}_{i=1}^{N},
\quad s_i\in \{1,\ldots,K\}\ \text{or}\ \Sigma.
\]

% ============================================================
\section{Transaction Construction and Rare Pattern Mining}

\subsection{Window-to-Transaction Mapping}

We slide a symbolic window of length $W$ tokens over $S(x)$.
Each symbolic window is transformed into a transaction:
\[
w_j = \{s_i : i\in \text{window }j\}.
\]

\textbf{Bigram enrichment.}
Converting a symbolic window to a set discards sequential order, which may cause the method
to miss ordering-sensitive anomalies (e.g., ST-elevation \emph{followed by} T-wave inversion
in ECG).
To partially recover ordering information while remaining within the itemset mining framework,
we augment each transaction with consecutive token bigrams:
\[
w_j^{+} = w_j \;\cup\; \{(s_i \!\to\! s_{i+1}) : i,i\!+\!1\in\text{window }j\}.
\]
For example, the token sequence $[A,B,C]$ yields the enriched transaction
$\{A,B,C,A{\to}B,B{\to}C\}$.
Bigrams add lightweight order sensitivity without requiring a full sequential pattern miner. The corresponding transaction database is:
\[
\Tcal = \{(w_j^{+},[\tau_j^s,\tau_j^e])\}_{j=1}^{M}.
\]
where ($\tau_j^s$,$\tau_j^e$) is the time interval spanned by the tokens inside transaction ($w_j^{+}$).
\subsection{Rare Itemsets and Rare Rules}

Let $I$ be an itemset.
Its support on the normal training transaction database is:
\[
\text{supp}(I)=
\frac{1}{|\Tcal_{\text{train}}|}
\sum_{w\in \Tcal_{\text{train}}}\mathbf{1}[I\subseteq w].
\]
An itemset is rare if $\text{supp}(I)\leq \tau$.
We denote the family of mined rare itemsets as $\Rcal$.

\textbf{Monotonicity of rarity.}
If $\text{supp}(I)\leq\tau$, then for any superset $I'\supset I$,
$\text{supp}(I')\leq\text{supp}(I)\leq\tau$.
Therefore the family of rare itemsets is \emph{upward-closed}, and \textbf{minimal rare
itemsets} (MRIs) serve as generators of the entire rare family.
Mining only MRIs avoids redundancy: we identify the boundary between frequent and rare
without enumerating exponentially many supersets \cite{Valtchev07}.

\textbf{Mining backend.}
We use a level-wise BtB (Break-the-Barrier) algorithm that extends Apriori with a
rarity boundary: frequent itemsets at level $k$ generate candidates at level $k\!+\!1$;
candidates whose support falls below $\tau$ are recorded as MRIs and excluded from further
expansion.
This yields the complete set of MRIs efficiently.

% ============================================================
\section{Rare-Pattern Anomaly Scoring}

\subsection{Window-Level Score}

For a transaction window $w_j$, we define:
\begin{equation}
S_{\text{win}}(w_j)=
\sum_{I\in \Rcal:\; I\subseteq w_j}\alpha(I),
\label{eq:win_score}
\end{equation}
where:
\begin{equation}
\alpha(I)=
\frac{g(|I|)}{\text{supp}(I)+\eps}.
\label{eq:item_weight}
\end{equation}
Here $g(|I|)=\log(1+|I|)$ and $\eps$ is a small constant for numerical stability.

\textbf{Information-theoretic interpretation.}
The weight $\alpha(I)$ is inspired by surprisal: rarer itemsets should receive larger weights. Instead of using the logarithmic term $-\log \mathrm{supp}(I)$, we use the inverse-frequency factor $1/(\mathrm{supp}(I)+\epsilon)$ as a stronger rarity amplifier, since it grows faster as support approaches zero. The length factor $g(|I|)=\log(1+|I|)$ mildly favors structured multi-token co-occurrences over rare singletons.

\subsection{Record-Level Score}

A record-level anomaly score aggregates the most suspicious windows:
\begin{equation}
S_{\text{rec}}(x)=
\operatorname{TopKMean}\left(\{S_{\text{win}}(w_j)\}_{j=1}^{M}\right).
\label{eq:rec_score}
\end{equation}
Top-$K$ aggregation is robust to isolated noise while preserving localized abnormal bursts.

\subsection{Interval Extraction}

Window scores are transformed into anomaly intervals through thresholding and merging:
\begin{enumerate}
  \item mark $w_j$ anomalous if $S_{\text{win}}(w_j)>\gamma$;
  \item merge consecutive anomalous windows within a gap tolerance;
  \item assign each interval a confidence using the max or mean score;
  \item attach top-$K$ triggered rare itemsets as local evidence.
\end{enumerate}

For each detected interval $a$, we define:
\[
\text{Evidence}(a)=
\operatorname{TopK}\{(I,\alpha(I)): I\in \Rcal,\; I\subseteq w_j,\; w_j\subseteq a\}.
\]

% ============================================================
\section{Allen Temporal Reasoning}

\subsection{Allen Relations}

Given two intervals $A=(a^s,a^e)$ and $B=(b^s,b^e)$, Allen interval algebra~\cite{allen1983}
defines 13 mutually exclusive qualitative relations: \texttt{before}, \texttt{after},
\texttt{meets}, \texttt{met-by}, \texttt{overlaps}, \texttt{overlapped-by}, \texttt{starts},
\texttt{started-by}, \texttt{during}, \texttt{contains}, \texttt{finishes},
\texttt{finished-by}, and \texttt{equals} (see Figure \ref{fig:pipeline_overview}, step 9).
We use these relations to describe how anomaly intervals interact with other anomalies,
artifacts, and contextual events.

\subsection{Composite Event Library}

We define a compact temporal rule library:
\begin{itemize}
  \item \textbf{Escalation:} repeated anomaly intervals that \texttt{meet} or strongly
  \texttt{overlap}, e.g., $A_1\ \texttt{meets}\ A_2\ \texttt{meets}\ A_3$.
  In ECG, this captures PVC couplets and triplets; in EEG, seizure buildup.
  \item \textbf{Artifact overlap:} an anomaly interval \texttt{overlaps} a known artifact
  interval, suggesting reduced detection confidence.
  \item \textbf{Staged onset:} an anomaly \texttt{starts} shortly after another event within
  a tolerance $\Delta t$, capturing pre-ictal to ictal transitions.
  \item \textbf{Cross-channel synchrony:} anomaly intervals on different channels
  \texttt{overlap} or \texttt{start} together, indicating a spatially distributed event.
\end{itemize}

\subsection{Temporal Relation Graph}

We build a graph $\Ecal=(V,E)$ where nodes are intervals and edges carry labeled Allen
relations:
\[
(a_i,\rho,a_j),\quad
\rho\in\{\texttt{meets},\texttt{overlaps},\ldots\}.
\]
Composite events are obtained by matching small patterns in this graph.

\textbf{Confidence refinement.}
Allen relations also enable post-hoc confidence adjustment: intervals participating in an
\texttt{artifact-overlap} relation have their confidence discounted, while intervals forming
an \texttt{escalation} chain have their confidence reinforced.
This mechanism provides a principled way for temporal reasoning to influence final anomaly
reporting beyond pure explanation.

% ============================================================
\section{Rare Temporal Concept Lattices}
\label{sec:fca}

This section introduces the main explanation layer of the paper.
Allen reasoning produces labeled temporal relations, but a record or cohort may still contain
many intervals and many local explanations.
To summarize them, we build an FCA formal context in which objects are detected anomaly
intervals and attributes describe their symbolic, temporal, and signal-context properties.

\subsection{Formal Context}

A formal context~\cite{ganter1999} is a triple: $\Kcal=(\Gcal,\Mcal,\mathcal{J})$, where $\Gcal$ is the set of detected anomaly intervals, $\Mcal$ is the set of binary
attributes, and $\mathcal{J}\subseteq \Gcal\times \Mcal$ is the incidence relation.
$(a,m)\in\mathcal{J}$ means that interval $a$ has property $m$.

\subsection{Attribute Construction}

We define five families of attributes.

\textbf{1.\;Rare symbolic attributes.}
For each rare itemset $I\in\Rcal$, we create:
$m_I = \operatorname{fires}(I)$.
An interval $a$ satisfies $m_I$ if at least one transaction inside $a$ contains $I$.

\textbf{2.\;Temporal attributes.}
For each Allen-derived composite, we create predicates such as:
$\operatorname{overlaps\mbox{-}artifact}$,
$\operatorname{meets\mbox{-}anomaly}$,
$\operatorname{escalation\mbox{-}member}$,
$\operatorname{starts\mbox{-}after\mbox{-}event}$.

\textbf{3.\;Signal-context attributes.}
These include channel/lead information and duration bins:
$\operatorname{lead\mbox{-}V5}$,
$\operatorname{frontal\mbox{-}EEG}$,
$\operatorname{duration\mbox{-}short}$,
$\operatorname{duration\mbox{-}long}$.

\textbf{4.\;Reconstruction fidelity attributes.}
When VQ-VAE tokenization is used, we record the mean reconstruction error per interval.
Intervals with error above the 90th percentile of normal training errors are assigned:
$\operatorname{high\mbox{-}recon\mbox{-}error}$.
This attribute flags intervals that are poorly represented by \emph{any} normal codebook
prototype, complementing the rare co-occurrence signal.

\textbf{5.\;Robustness attributes.}
During robustness tests, intervals may be labeled: \(\operatorname{stable\mbox{-}under\mbox{-}noise}\), \(\operatorname{stable\mbox{-}under\mbox{-}wander}\), or \(\operatorname{unstable\mbox{-}under\mbox{-}perturbation}\). Additional attributes can be added when further perturbations are evaluated.

\subsection{Formal Concepts}

For $A\subseteq \Gcal$, define
$A'=\{m\in\Mcal:\forall a\in A,\;(a,m)\in\mathcal{J}\}$.
For $B\subseteq\Mcal$, define
$B'=\{a\in\Gcal:\forall m\in B,\;(a,m)\in\mathcal{J}\}$.
A formal concept is a pair $(A,B)$ such that $A'=B$ and $B'=A$.
Here $A$ is the \textbf{extent} (anomaly intervals) and $B$ is the \textbf{intent}
(maximal shared properties).

Thus, a concept may be interpreted as:
\[
\underbrace{\{a_1,a_2,a_7\}}_{\text{similar anomalies}}
\;\Longleftrightarrow\;
\underbrace{
\{\operatorname{fires}(\{B_{42},B_{91}\}),\;
\operatorname{meets\mbox{-}anomaly},\;
\operatorname{lead\mbox{-}II}\}
}_{\text{shared explanation}}.
\]

\subsection{Toy Example}
To illustrate the construction, consider a simplified dataset with five anomaly intervals $\Gcal=\{A_1, \dots, A_5\}$ and five representative attributes $\Mcal=\{R_1, R_2, \text{long-duration}, \text{escalation}, \text{overlap-artifact}\}$, where $R_1$ and $R_2$ represent distinct rare symbolic patterns.
The formal context is represented as a binary cross-table (Table~\ref{tab:toy_context}). 

\begin{figure}[t]
\centering
\includegraphics[width=0.85\linewidth]{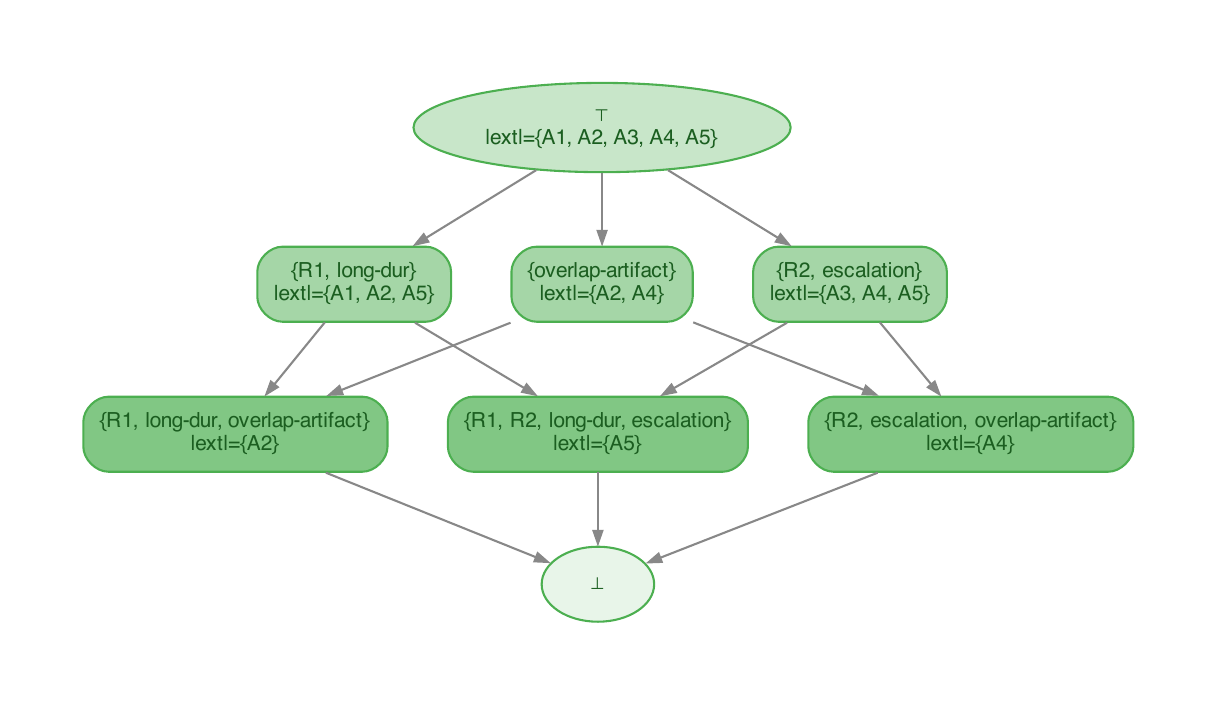}
\vspace{-2em}
\caption{Concept lattice of the toy formal context in Table~\ref{tab:toy_context}.}
\label{fig:toy_lattice}
\vspace{-1.2em}

\end{figure}

Applying FCA to this context generates the concept lattice shown in Figure~\ref{fig:toy_lattice}. The lattice organizes the intervals hierarchically based on shared attributes. The top node ($\top$) covers all intervals, while deeper nodes represent more specific concepts (more attributes, fewer intervals). For example, the concept $\{R_1, \text{long-duration}\}$ groups intervals $\{A_1, A_2, A_5\}$, while its sub-concept $\{R_1, \text{long-duration}, \text{overlap-artifact}\}$ specifically isolates $\{A_2\}$.
%\vspace{-2em}
\begin{table}[t]
\vspace{-1em}
\centering
\tiny
\caption{Toy Formal Context mapping anomaly intervals to attributes.}
\label{tab:toy_context}
\begin{tabular}{l|ccccc}
\toprule
Interval & $R_1$ & $R_2$ & long-duration & escalation & overlap-artifact \\
\midrule
$A_1$ & $\times$ & & $\times$ & & \\
$A_2$ & $\times$ & & $\times$ & & $\times$ \\
$A_3$ & & $\times$ & & $\times$ & \\
$A_4$ & & $\times$ & & $\times$ & $\times$ \\
$A_5$ & $\times$ & $\times$ & $\times$ & $\times$ & \\
\bottomrule
\end{tabular}
\vspace{-4em}
\end{table}

\subsection{Iceberg Lattice and Explanation Compression}

Full concept lattices may be exponentially large in $|\Mcal|$.
We therefore compute an \textbf{iceberg lattice} by retaining only concepts whose extent
satisfies $|A|\geq\sigma$, and restrict attributes to the top-$K$ most discriminative rare
patterns and the most informative Allen/context predicates.

\textbf{Complexity.}
Standard algorithms such as NextClosure~\cite{ganter1999} enumerate concepts in
$O(|\Gcal|\cdot|\Mcal|\cdot|\text{concepts}|)$ time.
The iceberg threshold $\sigma$ and attribute filtering to top-$K$ together bound the
effective number of concepts, keeping computation tractable for typical anomaly counts
(tens to low hundreds of intervals per record or cohort).

The concept lattice acts as an explanation-compression mechanism.
Instead of reporting all local interval explanations independently, we report concept
summaries:
$\text{Concept }c=(A_c,B_c):\;|A_c|\text{ intervals share }B_c$.

Algorithm~\ref{alg:signal2symbol} provides the full pseudo-code of the proposed Signal2Symbol pipeline.
\subsection{Example Interpretation}

A concept with intent
$B_c=\{\operatorname{fires}(\{B_{42},B_{91}\}),\;
\operatorname{overlaps\mbox{-}artifact},\;
\operatorname{lead\mbox{-}V5}\}$
may be summarized as:
\emph{``A family of V5 anomaly intervals is driven by the rare token co-occurrence
$\{B_{42},B_{91}\}$ but overlaps artifact intervals; these detections should be interpreted
with reduced confidence.''}

A concept with intent
$B_c=\{\operatorname{fires}(\{B_{17},B_{91}\}),\;
\operatorname{escalation\mbox{-}member},\;
\operatorname{duration\mbox{-}long}\}$
may be summarized as:
\emph{``A repeated long-duration anomaly type appears as an escalation chain, suggesting a
sustained arrhythmic episode or seizure buildup.''}

% ============================================================
\begin{algorithm}[t]
\caption{Signal2Symbol with Rare Temporal Concept Lattice}
\label{alg:signal2symbol}
\begin{algorithmic}[1]
\Require Signal record $x(t)$, tokenizer type, window length $W$, rarity threshold $\tau$,
  score threshold $\gamma$, support $\sigma$
\Ensure Record score, anomaly intervals, rare evidence, Allen graph, concept lattice
\State Segment $x(t)$ into windows or beats $\{x_i\}_{i=1}^{N}$
\State Convert segments into tokens via SAX or VQ-VAE
\State Build bigram-enriched transactions $\Tcal=\{w_j^{+}\}_{j=1}^{M}$
\State Mine MRI rare itemsets $\Rcal$ on normal training transactions
\For{each transaction $w_j^{+}$}
  \State Compute $S_{\text{win}}(w_j)$ via Eq.~\eqref{eq:win_score}
\EndFor
\State Aggregate $S_{\text{rec}}(x)$ via Eq.~\eqref{eq:rec_score}
\State Extract anomaly intervals $\Ical(x)$ 
\State Attach top-$K$ rare symbolic evidence to each interval
\State Compute Allen relations among anomaly and context intervals
\State Refine interval confidence via Allen-based adjustment
\State Construct FCA context $\Kcal=(\Gcal,\Mcal,\mathcal{J})$ with five attribute families
\State Generate iceberg concept lattice with support $\sigma$
\State \Return scores, intervals, evidence, Allen graph, concept lattice
\end{algorithmic}

\end{algorithm}

% ============================================================
\section{Experiments}
\label{sec:experiments}

\subsection{Datasets}

We evaluate on two ECG benchmarks and one EEG benchmark, chosen to cover beat-level,
record-level, and segment-level granularities across two signal modalities.

\begin{itemize}
  \item \textbf{MIT-BIH Arrhythmia Database}~\cite{mitbih}:
  48 half-hour, two-channel ambulatory ECG recordings at 360\,Hz with ${\sim}$110,000
  beat-level annotations spanning normal (N) and arrhythmic types (PVC, APC, LBBB, RBBB,
  Paced).
  We segment by R-peak and use beat-centered windows.
  Normal beats form the training set; arrhythmic beats are anomalies.
  Patient-wise splits prevent leakage.
  This dataset is our primary benchmark: beat-level annotations permit full evaluation of
  interval detection, Allen reasoning, and FCA concept formation across distinct arrhythmia
  families. Here, each beat is represented by a 216-sample R-peak-centered window, corresponding to 600~ms at 360~Hz.

  \item \textbf{PTB-XL}~\cite{ptbxl}:
  21,837 twelve-lead, 10-second ECG records at 100\,Hz with record-level diagnostic labels
  (NORM, MI, STTC, CD, HYP).
  We use NORM records for training and pathological records as anomalies.
  Patient-wise stratified splits follow the recommended folds.
  PTB-XL tests scalability to a large modern benchmark at record-level granularity. Here, we use the complete 10-s lead-II record, with 1,000 samples at 100~Hz.

  \item \textbf{Bonn EEG Dataset}~\cite{bonn_eeg}:
  500 single-channel EEG segments (23.6\,s each, 173.61\,Hz) across five classes:
  eyes-open (A), eyes-closed (B), interictal hippocampal (C), interictal epileptogenic (D),
  and ictal/seizure (E).
  We use classes A--B as normal and C--E as varying degrees of abnormality. Here, each recording is divided into 512-sample windows, corresponding to approximately 2.95~s at 173.61~Hz, with $50\%$ overlap. 
\end{itemize}

\subsection{Baselines}
We compare five scoring backbones: \textbf{IF} (Isolation Forest, 200 trees) and \textbf{LOF} ($k\!=\!20$) applied to flattened time-domain features; \textbf{USAD}~\cite{usad}, a deep learning baseline of two adversarially-trained autoencoders whose combined reconstruction error serves as the anomaly score; \textbf{SAX+RareScore}, using SAX tokens with rare-pattern scoring; and \textbf{Signal2Symbol} (VQ-VAE+RareScore), our full neuro-symbolic backbone. Detection metrics are reported for all scoring backbones; robustness is evaluated under signal perturbations, while Allen/FCA are assessed through explanation-compression and case-study analyses.

%We compare four scoring backbones: IF, LOF, SAX+RareScore, and the Signal2Symbol VQ-VAE+RareScore backbone. Detection metrics are reported for the scoring backbones; robustness is evaluated under signal perturbations, while Allen/FCA are assessed through explanation-compression and case-study analyses.
%\begin{itemize}
%
%  \item \textbf{IF / LOF:} Isolation Forest or Local Outlier Factor on time-domain
%  and frequency-domain features;
 % \item \textbf{SAX+RareScore:} SAX tokens with rare-pattern scoring;
 % \item \textbf{Signal2Symbol scoring backbone:} VQ-VAE tokens with rare-pattern scoring;
 %
%\end{itemize}
%Detection metrics are reported for the scoring backbones, while Allen/FCA are assessed through robustness and explanation-compression metrics.
\subsection{Metrics}

For detection we report AUROC and AUPRC. Explanation quality is assessed via: (i)~\emph{rule sparsity} (average evidence patterns per interval); (ii)~\emph{concept compression ratio} (local explanations divided by retained lattice concepts); (iii)~\emph{concept stability} (Jaccard overlap of concept intents across bootstrap samples); and (iv)~\emph{mean intent length} (average attributes per concept). Robustness is measured as $\Delta$AUROC under Gaussian noise injection (SNR $\in\{5,10,20\}$\,dB) and sinusoidal baseline wander (amplitudes $\in\{0.1,0.3,0.5\}$\,mV).

\subsection{Implementation Details}
The VQ-VAE encoder consists of three Conv1d layers (32, 64, 128 filters) with a linear projection to $\R^{64}$; the codebook ($K\!=\!128$) uses EMA updates with commitment weight $\beta\!=\!0.25$ and is trained for 100 epochs on normal segments with MSE loss. USAD shares the same encoder architecture with two decoders trained adversarially for 100 epochs. For SAX, we use alphabet size $A\!=\!5$ and PAA reduction to 16 segments per window. The rare itemset miner uses $\tau\!=\!0.02$ and max itemset length 5, with bigram-enriched transactions of window size $W\!=\!8$ and step size 1. Anomaly scoring uses $\eps\!=\!10^{-6}$, Top-$K$ aggregation ($K\!=\!5$), and score threshold $\gamma$ at the 95th percentile of normal validation scores. The FCA iceberg lattice retains concepts with support $\sigma\!\geq\!2$ and top-50 attributes.

\subsection{Main Detection Results}

Table~\ref{tab:main_results} reports detection performance across all three datasets. The results show that learned symbolic tokenization consistently improves rare-pattern anomaly ranking over SAX. VQ-VAE+Rare achieves the best AUROC and AUPRC on all three datasets, with particularly large gains on PTB-XL and MIT-BIH, suggesting that learned codebook prototypes capture morphology-level variations that fixed SAX discretization misses. On Bonn EEG, all symbolic methods obtain high AUPRC, while VQ-VAE+Rare further improves AUROC, indicating better separation between normal and abnormal EEG segments. IF and LOF provide competitive classical baselines, while USAD offers a deep reconstruction baseline; nevertheless, Signal2Symbol achieves the best AUROC and AUPRC across all three datasets. These results support the use of VQ-VAE tokens as a stronger symbolic substrate for the subsequent Allen/FCA explanation layers.

\begin{table}[t]
\tiny
\centering
\caption{Detection results across ECG and EEG datasets.
MIT-BIH is evaluated at beat-level, PTB-XL at record-level (lead~II only), and Bonn at segment-level.}
\label{tab:main_results}
\begin{tabular}{l l cc}
\toprule
Dataset & Method & AUROC$\uparrow$ & AUPRC$\uparrow$ \\
\midrule
\multirow{4}{*}{\rotatebox{90}{MIT-BIH}}
& IF (200 trees)       & 0.790 & 0.405 \\
& LOF ($k\!=\!20$)     & 0.737 & 0.423 \\
&USAD &0.750& 0.341\\
& SAX+Rare             & 0.605 & 0.198 \\
& Signal2Symbol (VQ-VAE+Rare)         & \textbf{0.818} &\textbf{ 0.547} \\
\midrule
\multirow{3}{*}{\rotatebox{90}{ PTB-XL}}
& IF (200 trees)       & 0.546 & 0.615 \\
& LOF ($k\!=\!20$)     & 0.517 & 0.592 \\
&USAD &0.516&0.588 \\
& SAX+Rare             & 0.493 & 0.574 \\
& Signal2Symbol  (VQ-VAE+Rare)        & \textbf{0.742} & \textbf{0.634} \\
\midrule
\multirow{3}{*}{\rotatebox{90}{ Bonn}}
& IF (200 trees)       & 0.469 & 0.876 \\
& LOF ($k\!=\!20$)     & 0.512 & 0.896 \\
&USAD &0.550& 0.900\\
& SAX+Rare             & 0.579 & 0.911 \\
& Signal2Symbol   (VQ-VAE+Rare)       & \textbf{0.710} & \textbf{0.959} \\
\bottomrule
\end{tabular}
\vspace{-1em}
\end{table}

\subsection{Ablation Study}

Table~\ref{tab:ablation} ablates key design choices on the Bonn EEG dataset. The ablation confirms that learned VQ-VAE tokens provide a stronger symbolic representation than SAX on Bonn EEG, with performance saturating already at $K=128$. Increasing the SAX alphabet or PAA resolution does not improve detection, suggesting that finer fixed discretization may fragment useful motifs. %The support-threshold results further show that mining too many MRIs can hurt precision, while a higher threshold yields fewer, more stable rare patterns.

\begin{table}[t]
\centering
\tiny
\caption{Ablation study: effect of tokenizer, alphabet size,
and PAA resolution on Bonn EEG.}
\label{tab:ablation}
\begin{tabular}{lccl}
\toprule
Setting & AUROC$\uparrow$ & AUPRC$\uparrow$ & Notes \\
\midrule
\multicolumn{4}{l}{\textit{Tokenizer}} \\
SAX ($|\Sigma|\!=\!5, P\!=\!16$)  & 0.579 & 0.911 & fixed discretization \\
SAX ($|\Sigma|\!=\!6, P\!=\!16$)  & 0.536 & 0.898 & larger alphabet \\
SAX ($|\Sigma|\!=\!6, P\!=\!20$)  & 0.512 & 0.891 & more segments \\
VQ-VAE ($K\!=\!128$)   & \textbf{0.710} & \textbf{0.959} & learned codebook \\
VQ-VAE ($K\!=\!256$)   & \textbf{0.710} & \textbf{0.959} & saturates at $K\!=\!128$ \\
%\midrule
%\multicolumn{4}{l}{\textit{Support threshold $\tau$ (SAX, $|\Sigma|\!=\!5$)}} \\
%$\tau_{\min} = 0.02$   & 0.605 & 0.198 & more MRIs (1448) \\
%$\tau_{\min} = 0.05$   & 0.579 & 0.911 & fewer MRIs (409) \\
\bottomrule
\end{tabular}
\vspace{-2em}
\end{table}

\subsection{Sensitivity to Support Threshold}

\begin{figure}[ht]
\centering
\includegraphics[width=0.8\linewidth]{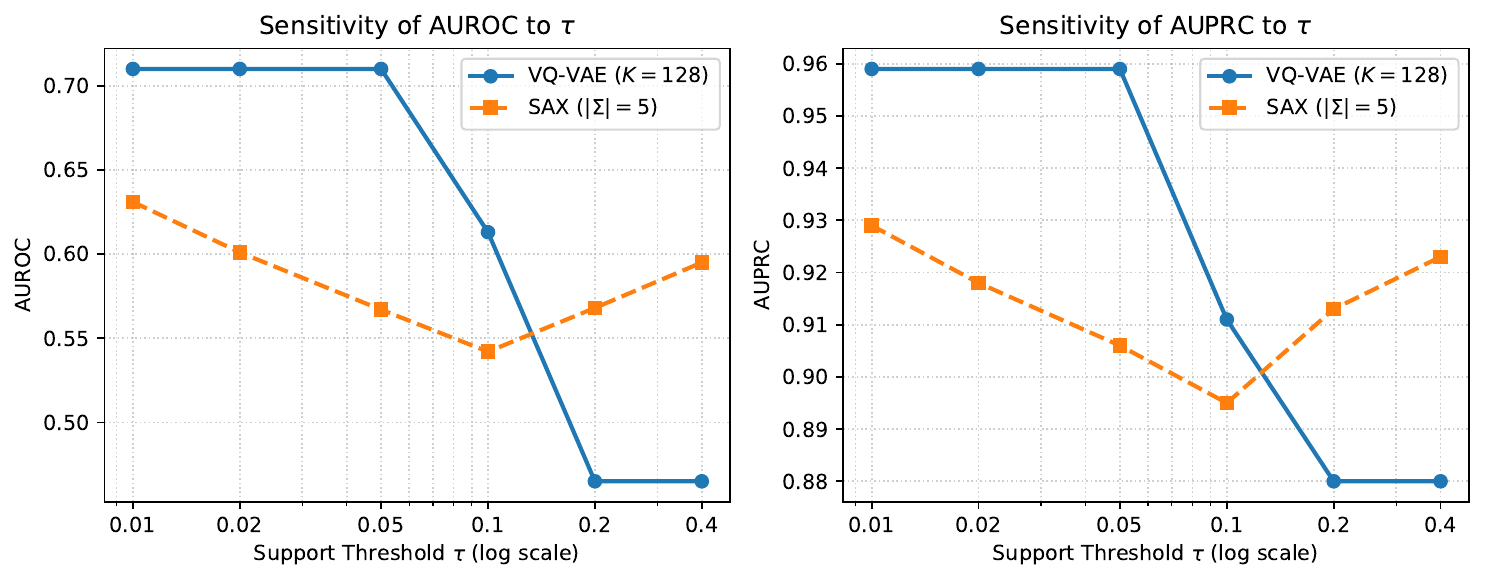}
\caption{Sensitivity of AUROC and AUPRC to the support threshold $\tau$ on the Bonn EEG dataset. The VQ-VAE model exhibits remarkable stability across lower thresholds.}
\label{fig:sensitivity}
\vspace{-0.7em}
\end{figure}

The pipeline involves interacting thresholds, notably the rarity threshold $\tau$ used during mining. To evaluate the robustness of our models to this hyperparameter, we performed a grid search over $\tau \in [0.01, 0.40]$ on the Bonn EEG dataset (Figure~\ref{fig:sensitivity}).
The VQ-VAE ($K\!=\!128$) model demonstrates remarkable stability, maintaining its peak AUROC of 0.710 across all $\tau\leq 0.05$. This occurs because the model extracts a highly specific, robust set of 21 Minimal Rare Itemsets (MRIs) that capture the dominant anomalous symbolic configurations. In contrast, the SAX tokenizer is much more sensitive to thresholding, requiring the mining of over 4,000 dense MRIs at $\tau\!=\!0.01$ to marginally improve from 0.601 to 0.631 AUROC, sacrificing explainability. These results confirm that our neuro-symbolic approach is highly robust to hyperparameter selection and does not require aggressive grid searching to achieve peak anomaly detection performance. Preliminary sensitivity checks on the remaining datasets showed the same trend.
\vspace{-0.3em}
%%%%%%%
\subsection{Explanation Quality}
Table~\ref{tab:explain} reports explanation metrics on Bonn EEG, PTB-XL and MIT-BIH.
\begin{table}[t]
\tiny
\centering
\caption{Explanation quality across datasets. Compression is
the ratio of detected anomaly intervals to retained iceberg-lattice
concepts; Intent is the mean number of attributes per retained
concept.}
\label{tab:explain}
\begin{tabular}{lcccc}
\toprule
Dataset & Intervals & Concepts & Compression$\uparrow$ & Intent$\downarrow$ \\
\midrule
Bonn EEG & 2061 & 40 & \textbf{51.5$\times$} & \textbf{3.2} \\
MIT-BIH  & 2154 & 32 & \textbf{67.3$\times$} & \textbf{4.1} \\
PTB-XL   & 585  & 144 & \textbf{4.1$\times$}  & \textbf{4.0} \\
\bottomrule
\end{tabular}
\vspace{-2em}
\end{table}
As illustrated, our full pipeline achieves significant explanation compression across all three primary datasets. For Bonn EEG, the FCA layer systematically groups 2,061 individual anomaly intervals into just 40 distinct formal concepts, achieving a $51.5\times$ compression ratio. On the much larger MIT-BIH dataset, the compression is even stronger at $67.3\times$, collapsing 2,154 raw anomaly intervals into 32 high-level prototypes. Even on PTB-XL's complex, multi-pathology ECG records, the lattice compresses 585 raw anomalous intervals down to 144 formal concepts (a $4.1\times$ reduction). This allows domain experts to review a manageable handful of anomaly prototypes rather than thousands of isolated events. Furthermore, the average concept intent across all datasets consists of roughly 3 to 4 attributes, meaning each anomaly group is concisely described by a small, interpretable set of temporal and symbolic properties.
\subsection{Robustness}

Table~\ref{tab:robustness} reports $\Delta$AUROC under perturbation on Bonn EEG. The VQ-VAE tokenizer shows no observable AUROC degradation under the tested perturbation settings, maintaining the same measured AUROC of 0.710 within the reported precision ($\Delta = 0.000$) across all tested noise levels (up to 5~dB SNR) and baseline wander amplitudes. In contrast, the SAX baseline suffers minor degradation under noise and a more noticeable performance drop ($\Delta = -0.024$) when subjected to baseline wander. The neural codebook's ability to smoothly map corrupted continuous segments to the same discrete latent prototypes makes it highly resilient to common clinical artifacts.

\begin{table}[t]
\centering
\tiny
\caption{Robustness ($\Delta$AUROC vs.\ clean) on Bonn EEG.
Noise column reports worst-case $\Delta$ across SNR $\in \{5,10,20\}$~dB;
Wander column across amplitudes $\in \{0.1,0.3,0.5\}$.}
\label{tab:robustness}
\begin{tabular}{lccc}
\toprule
Method & Clean AUROC & Noise $\Delta$ & Wander $\Delta$ \\
\midrule
SAX+Rare ($|\Sigma|\!=\!5$) & 0.579 & $-$0.002 & $-$0.024 \\
VQ-VAE+Rare ($K\!=\!128$) & \textbf{0.710} & \textbf{0.000} & \textbf{0.000} \\
\bottomrule
\end{tabular}
\vspace{-2em}
\end{table}
\subsection{Case Study: Explainability via Concept Lattices}
\label{sec:case_study}

\begin{figure*}[t]
\centering
\includegraphics[width=\linewidth]{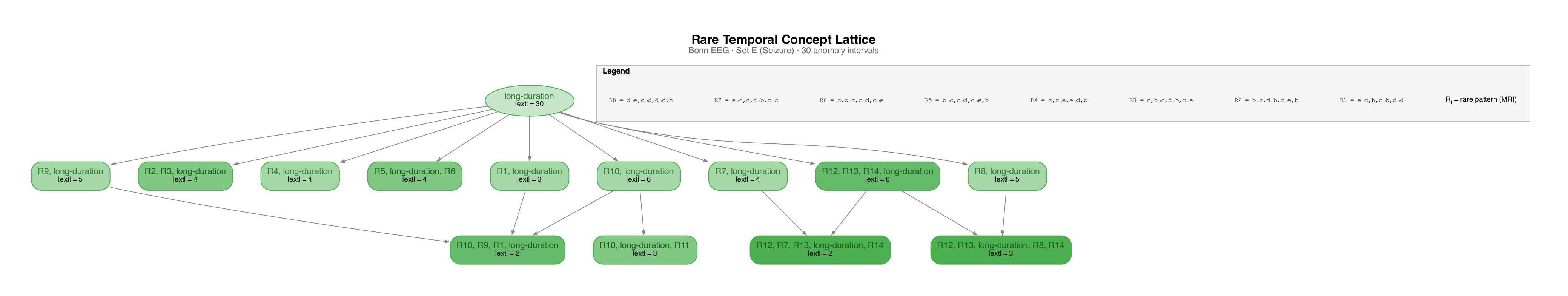}
\vspace{-2em}
\caption{Rare Temporal Concept Lattice generated from 30 anomaly intervals on Bonn EEG (Set E, Seizure). The FCA lattice compresses individual detections into a hierarchy of concepts based on shared temporal attributes (e.g., \emph{long-duration}) and rare symbolic patterns ($R_i$). Deeper nodes represent more specific anomaly signatures.}
\label{fig:fca_lattice}
\vspace{-1.5em}
\end{figure*}

We present a qualitative case study on the Bonn EEG dataset (Set E, seizure recordings) to demonstrate how Formal Concept Analysis (FCA) provides structured explainability.
Figure~\ref{fig:fca_lattice} visualizes a concept lattice built from 30 detected anomaly intervals during a seizure event.

\textbf{Local Symbolic Evidence.} 
The symbolic tokenizer and rare pattern miner identify specific rare transitions and co-occurrences (e.g., $R_1, R_2$) that flag these windows as anomalous against the background normal activity.

\textbf{Temporal Reasoning.} 
The Allen temporal graph connects these intervals. The reasoner identifies that these anomalies are contiguous and overlapping, forming extended periods of abnormal activity, leading to the \emph{long-duration} attribute.

\textbf{Concept Lattice.} 
Rather than presenting the clinician with 30 independent alarms, the FCA lattice (Figure~\ref{fig:fca_lattice}) organizes them hierarchically. 
The top concept ($\top$) encompasses all 30 intervals, establishing that they all share the \emph{long-duration} temporal characteristic.
Sub-concepts then branch out based on specific subsets of rare patterns that fired within those intervals.
For instance, the lattice groups sets of intervals that co-exhibit specific rare patterns, effectively compressing 30 distinct anomalies into 14 interpretable, rule-based families.
% ============================================================
\section{Discussion}

Signal2Symbol is designed to make physiological anomaly detection explainable at three levels.
First, rare itemsets provide \textbf{local symbolic evidence}: they identify the unusual
token co-occurrences that caused a window to be scored as anomalous.
Second, Allen relations provide \textbf{temporal structure}: they describe how intervals
meet, overlap, or follow one another, and enable confidence refinement.
Third, FCA provides \textbf{conceptual compression}: it groups intervals into shared
temporal-symbolic anomaly families. This combination is important because physiological time series often contain many local
irregularities, not all of which are clinically meaningful.
A list of isolated detections is difficult to interpret.
A concept lattice, however, can reveal that several intervals share the same rare motif and
the same temporal behavior.
For example, artifact-overlap concepts can be separated from escalation concepts, and
cross-channel concepts can be distinguished from single-channel events.

\textbf{Why VQ-VAE helps.}
SAX is transparent and efficient, but it discretizes mostly amplitude patterns.
VQ-VAE can learn recurring waveform morphologies, making token co-occurrences more
semantically meaningful.
This is especially relevant for ECG beat morphology (e.g., distinguishing PVC from LBBB
shapes) and EEG transient patterns.

\textbf{Why bigrams help.}
Pure set-valued transactions lose all sequential structure.
Bigram enrichment recovers pairwise ordering at negligible cost, enabling the detection of anomalies that depend on transition patterns (e.g., normal-to-PVC transitions)
without requiring a full sequential pattern miner.

\textbf{Why FCA helps.}
Rare patterns alone may produce many fragmented explanations.
FCA organizes them into a lattice of shared attributes, enabling visual inspection and
explanation compression.
This makes the method suitable as an AI reasoning and knowledge-discovery tool.

\textbf{Hyperparameter sensitivity.}
The pipeline involves several interacting thresholds: rarity threshold $\tau$, window size
$W$, codebook size $K$, score threshold $\gamma$, and concept support $\sigma$.
We note a qualitative interaction: larger $W$ increases token co-occurrence within
transactions, raising support values and requiring a correspondingly higher $\tau$ to
identify rare patterns.
The ablation study varies these parameters individually; a full factorial analysis is left
to future work.

\textbf{Limitations.}
First, bigram enrichment recovers only pairwise order; longer sequential dependencies
require n-gram or sequential rule mining.
Second, the lattice can become large if too many attributes are included; we address
this with iceberg support thresholds and top-$K$ attribute filtering.
Third, VQ-VAE may suffer from dead codes without careful training (EMA updates and
utilization tracking help).
Fourth, the Allen/FCA explanation quality depends on reliable interval extraction; noisy
thresholds can fragment or merge intervals incorrectly.

\textbf{Per-dataset analysis.}
On the Bonn EEG dataset, Signal2Symbol (VQ-VAE+Rare) achieves 0.71 AUROC, substantially outperforming
Isolation Forest (0.46) and SAX+Rare (0.57). EEG seizure activity produces distinctive
temporal patterns that are well captured by learned codebook prototypes.
On MIT-BIH, IF operates directly on the 216-dimensional beat morphology and achieves
0.79 AUROC, while our symbolic methods reach 0.81 (VQ-VAE) and 0.60 (SAX). 
On PTB-XL, IF achieves near-random performance (0.54 AUROC), while our approach reaches a score of 0.74. 

\section{Conclusion and Future Work}

We presented Signal2Symbol, a neuro-symbolic pipeline for explainable ECG/EEG anomaly
detection.
The method combines VQ-VAE/SAX tokenization, bigram-enriched rare-pattern anomaly scoring,
Allen interval reasoning with confidence refinement, and FCA-based rare temporal concept
lattices.
Evaluation on MIT-BIH, PTB-XL, and Bonn EEG demonstrates the framework across beat-level,
record-level, and segment-level granularities in two signal modalities.
The resulting framework produces not only anomaly scores, but also symbolic evidence,
temporal relations, and concept-level visual explanations. Future work will extend the current framework with order-aware sequence mining (n-grams and
sequential rules), masked token modeling over learned biosignal tokens, larger cross-dataset
evaluation including CHB-MIT seizure detection, and human-in-the-loop assessment of
explanation usefulness in clinical workflows.
\section*{Declarations}
The authors declare no competing interests. Public datasets were used; code and processed data are available upon request.% ============================================================

\end{document}